\documentclass[letterpaper, 10pt, twocolumn]{article}
\usepackage{clean}
\shorttitle{Visio-Verbal Teleimpedance Interface}

\usepackage{stfloats}
\usepackage{verbatim}
\usepackage{graphicx}

\usepackage{cite}
\usepackage{graphics}
\usepackage{epsfig}
\usepackage{bm,times}
\usepackage{amsmath}
\usepackage{amssymb}
\usepackage{amsfonts}

\usepackage{epstopdf}
\usepackage{float}
\usepackage{color}
\usepackage{url}
\usepackage{enumerate}
\usepackage{hyperref}

\graphicspath{{figs/}}

\begin{document}

\title{
Visio-Verbal Teleimpedance Interface: Enabling Semi-Autonomous Control of Physical Interaction via Eye Tracking and Speech
}

\author{
    Henk H.A. Jekel$^{1}$,
    Alejandro D\'{i}az Rosales$^{1,2}$, and
    Luka Peternel$^{1}$
    \thanks{
        $^{1}$Department of Cognitive Robotics, Delft University of Technology, Mekelweg 2, 2628 CD Delft, The Netherlands.}
    \thanks{
        $^{2}$European Organization for Nuclear Research (CERN), Espl. des Particules 1, 1211 Meyrin, Switzerland.}
    \thanks{
        E-mail: {\tt\small l.peternel@tudelft.nl}}
}

\maketitle

\begin{abstract}: 
The paper presents a visio-verbal teleimpedance interface for commanding 3D stiffness ellipsoids to the remote robot with a combination of the operator's gaze and verbal interaction. The gaze is detected by an eye-tracker, allowing the system to understand the context in terms of what the operator is currently looking at in the scene. Along with verbal interaction, a Visual Language Model (VLM) processes this information, enabling the operator to communicate their intended action or provide corrections. Based on these inputs, the interface can then generate appropriate stiffness matrices for different physical interaction actions. To validate the proposed visio-verbal teleimpedance interface, we conducted a series of experiments on a setup including a Force Dimension Sigma.7 haptic device to control the motion of the remote Kuka LBR iiwa robotic arm. The human operator's gaze is tracked by Tobii Pro Glasses 2, while human verbal commands are processed by a VLM using GPT-4o. The first experiment explored the optimal prompt configuration for the interface. The second and third experiments demonstrated different functionalities of the interface on a slide-in-the-groove task.
\end{abstract}


\section{Introduction}
\label{sec:introduction}
Teleoperation is a key technology enabling remote human control and teaching of robots in scenarios such as disaster response, robot-assisted surgery, inspection \& maintenance, space exploration, and hazardous environment operations~\cite{si2021review}. While autonomous robots excel in structured environments like manufacturing, they struggle to adapt to dynamic, unstructured conditions due to limited cognitive flexibility. In that respect, teleoperation offers more adaptability by integrating humans into the robot control loop, allowing operators to issue motion commands through interfaces such as haptic devices, joysticks, and motion capture systems.

Nevertheless, controlling the motion alone makes the execution of complex tasks in interaction with unstructured and unpredictable environments difficult~\cite{suomalainen2022survey}. Impedance control enables the control of the relationship between forces and motion and simplifies the execution of tasks with physical interactions~\cite{naceri2021learning}. In this case, stiffness is the most important parameter of impedance as it controls how soft or stiff the robot is when interacting with fragile objects. Teleimpedance is a concept that allows human operators to control the stiffness of the remote robot through various interfaces~\cite{peternel2022after}.

\begin{figure}[t]
    \centering
    \includegraphics[width=1\linewidth]{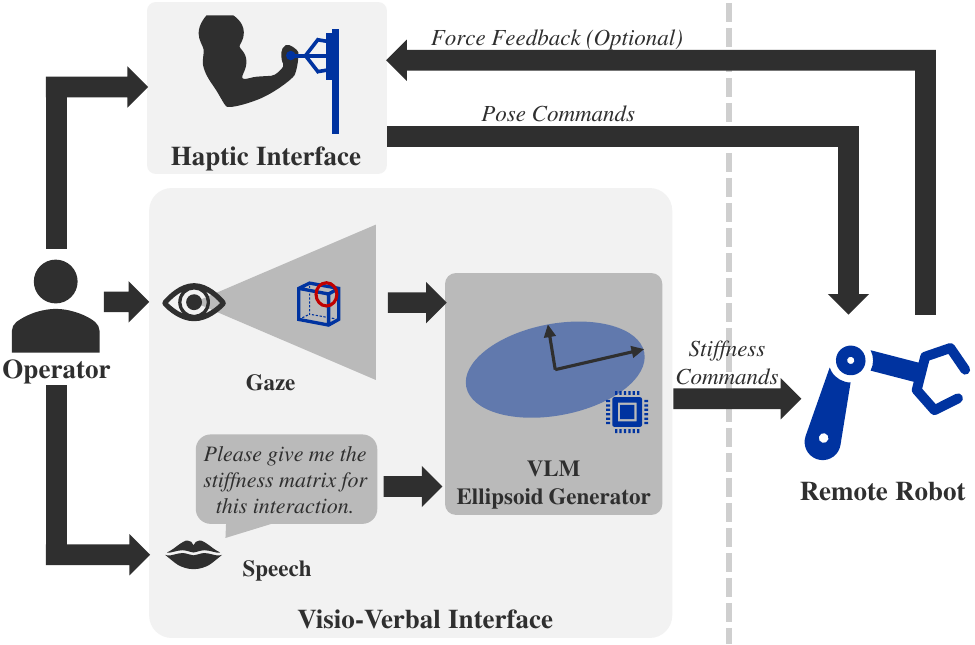}
    \caption{Diagram presenting the visio-verbal teleimpedance interface inside a teleimpedance setup. The local and remote sites are separated by the dotted line. At the local site, the operator controls the remote robot using a haptic device to command motion. The visio-verbal interface operates in parallel, allowing the operator to configure the desired stiffness using gaze and speech inputs. This stiffness configuration, along with motion commands, is then transmitted to the remote robot.}
    \label{fig:interface_setup}
\end{figure}

Existing teleimpedance command interfaces can be broadly categorized into manual impedance control and automated impedance control, depending on whether the human or the robot automation determines the impedance of the interaction~\cite{peternel2022after}. Manual control approaches allow the human operator to directly adjust stiffness through different inputs. One common method involves estimating the operator’s stiffness with muscle activity using electromyography (EMG), which is then mapped to the robot’s impedance~\cite{ajoudani2012tele,yang2018dmps,park2018programming,buscaglione2022tele}. While EMG-based teleimpedance interfaces enable intuitive multi-degree of freedom (DoF) control of impedance, they require sensor placement and calibration procedures and have limited generalisability. Simplification in terms of the number of EMG sensors alleviates some of these issues, however, they require an additional motion capture system to retain some of multi-DoF functionality~\cite{ajoudani2018reduced}. Interfaces based on biosignals are also subject to a neuromechanical coupling effect between the commanded stiffness and force feedback, which takes away some of the control due to reflexes~\cite{doornebosch2021analysis}.

More practical teleimpedance interfaces are based on controlling stiffness with force grip sensor~\cite{walker2011demonstrating} and buttons~\cite{garate2021scalable}, joysticks/scroll-wheels~\cite{kraakman2024design}, perturbations on haptic device~\cite{gourmelen2021human}, tablets~\cite{peternel2021independently}, or augmented-reality~\cite{diazrosales2024interactive}. Nevertheless, they either have limited control over the stiffness ellipsoid in terms of DoF, perturb the operator, or take the operator's attention away from the task and impose a high cognitive workload on the operator.

In contrast to manual impedance control, automated impedance control systems remove the operator from the impedance control loop, allowing robot autonomy to determine the appropriate stiffness. For example, in the method in~\cite{michel2021bilateral}, the robot used torque sensors to measure physical interaction forces and autonomously adjust stiffness for task stabilization, ensuring compliance when needed and increasing rigidity during disturbances. Similarly, the work in~\cite{siegemund2024semi} proposed a vision-based system where the robot detects material properties and geometry of an object that the operator was about to touch with the remote robot. Based on that, the robot autonomy preemptively sets optimal stiffness values for the expected physical interaction. While these autonomous approaches improve safety and reduce operator workload, they remove direct human input and reduce the operator’s ability to intervene.

Despite advancements in teleimpedance interfaces, current methods either require continuous manual adjustments from the operator, which are cognitively demanding or rely on automation, which limits human input. To bridge this gap, this work introduces a novel visio-verbal teleimpedance interface that follows the shared control paradigm. By leveraging gaze and speech, two natural and intuitive communication modalities, this interface allows operators to configure the robot’s 3D stiffness ellipsoid, without diverting their visual attention from the task. The operators can inform the robot autonomy what they intend to do by conversation and gaze, and if the decisions of autonomy are unsatisfactory, corrections can be communicated. The conversation could be conducted by asking the operator to write down the request. However, during a study in~\cite{diazrosales2024interactive}, it was noted that the expert operator participant indicated a preference to control the stiffness using voice commands to keep their hands on the robot's movement controls. As a result, speech recognition modality was explored for the interface proposed in this paper.

Unlike previous methods, which either require direct manual adjustments or rely entirely on automation, this approach distributes cognitive workload between the human operator and a vision-language model (VLM). The interface provides gaze-based contextual awareness and interprets verbal commands to dynamically generate a stiffness matrix that optimizes the robot’s interaction with its environment. One of the key contributions of this work is to provide the first teleimpedance interface that combines gaze and speech modalities.

The rapid advancements in Deep Learning, particularly the transformer architecture~\cite{vaswani2017attention} and the increasingly larger model sizes~\cite{kaplan2020larger_model_size_is_better} have enabled breakthrough applications in speech-to-text, text-to-speech, and natural language processing combined with computer vision through VLMs. A popular example is ChatGPT, which uses a VLM to process both textual and visual inputs. While these models can generate highly contextual responses, they typically lack direct awareness of user focus unless extensively prompted. Recently, researchers proposed GazeGPT, which incorporates mobile eye-tracking to enhance context awareness by identifying where a user is looking~\cite{konrad2024gazegpt}. Building upon these advancements, we develop VLM-driven teleimpedance control integrating eye tracking into a visio-verbal system.

In the experimental work, we first conducted an extensive prompt parameter optimization study, specific to the use of VLMs in teleimpedance, where we investigated and compared the effects of different settings (system role, amount of priors, and image resolution) on the stiffness matrix predictions. To demonstrate and validate the key functionalities of the developed novel visio-verbal teleimpedance interface, we conducted experiments on a setup where Force Dimension Sigma.7 haptic device controls the motion of the remote Kuka LBR iiwa robotic arm. The human operator's gaze is tracked by Tobii Pro Glasses 2, while human verbal commands are processed by a VLM using GPT-4o. Two scenarios were investigated: the streamlined verbal mode of the interface, demonstrating fluid task execution, and the full mode, demonstrating the combined gaze and conversational aspects. The main contributions of this paper are: 1) the first-ever visio-verbal teleimpedance interface combining VLMs and gaze tracking, and 2) new insights about the effects of different settings on the stiffness matrix predictions in teleimpedance.

\section{Methods}
\label{sec:methods}
The concept of the proposed approach is illustrated by Fig.~\ref{fig:interface_setup}, where gaze and speech are used to determine the appropriate stiffness configuration for the remote robot. This is done in a natural manner without diverting their visual attention from the teleoperation task. Through the operator's speech and gaze inputs, the VLM ellipsoid generator defines the stiffness ellipsoid for the given physical interaction. The visio-verbal teleimpedance interface was designed to enable hands-free adjustment of the orientation and shape of the 3D translational stiffness ellipsoid. This means that the human operator can use their hands exclusively to control the motion of the remote robot in real-time through a haptic device.

To achieve the stiffness generation, the interface captures a snapshot of the teleoperation scene after the operator says ``capture'', overlaying a red circle to indicate the operator’s gaze estimate. Afterwards, the system processes the operator’s verbal commands regarding their intention. The verbal interactions are processed through a speech-to-text module. The VLM then interprets these multimodal inputs to generate an updated stiffness matrix, which is applied directly to the remote robot. The advantage of additional gaze input to the VLM is to reduce the amount of speech needed to indicate the operator's intention. To facilitate verification, the system provides immediate feedback through both verbal confirmation and real-time visualization of the updated stiffness ellipsoid, ensuring that adjustments align with the operator’s intent.

\subsection{Design Requirements}
\label{subsec:visio_verbal_teleimpedance_interface_design}
Based on the discussed aspects in the introduction and analysis of related work~\cite{peternel2022after}, we formulated a set of design requirements to guide the development of the novel \emph{visio-verbal} teleimpedance command interface. These requirements aim to address existing challenges and prioritize a human-centric design. The requirements are as follows:

\begin{enumerate}[ R1:]
    \item Combine gaze tracking with verbal interaction for robot stiffness matrix generation.\label{Requirements:1}
    \item Enable control of full 3D stiffness ellipsoid.\label{Requirements:2}
    \item Minimize visual distractions to ensure the operator's uninterrupted focus on the task.\label{Requirements:3}
    \item Minimal setting up and calibration procedures.\label{Requirements:4}
    \item Avoid the neuromechanical coupling effect between the commended stiffness and force feedback.\label{Requirements:5}
\end{enumerate}

To incorporate R\ref{Requirements:1}, the system follows a human-in-the-loop shared control paradigm, where the VLM assists in shaping and directing the stiffness ellipsoid, while still allowing the operator to refine its final configuration via voice commands. The system integrates a VLM to interpret high-level semantic inputs and gaze-based contextual awareness to quickly interpret what the operator is talking about in the remote scene. This provides a good middle ground between manual and autonomous variable impedance control. Using interpreted context from the speech and gaze, the system can generate a 3D stiffness ellipsoid to fit the given interaction in a specific task phase, thus achieving R\ref{Requirements:2}.

The combination of gaze and speech inputs with real-time multimodal processing simplifies stiffness modulation by eliminating the need for physical input devices. For instance, the teleimpedance interface that forms ellipsoids on a tablet allows for the generation of a 3D stiffness ellipsoid. However, this approach can distract the operator from the task at hand. In contrast, the proposed interface enables the operator to maintain focus on the task while an eye tracker monitors their gaze in the remote scene, thus achieving R\ref{Requirements:3}.

Since the microphone and eye tracker hardware are easy to use, the setup procedure is very short. Similarly, the calibration for the eye tracker takes only a minute. Therefore, avoiding the use of biosignals, we address R\ref{Requirements:4}. Finally, since the voice and gaze are both non-physical variables, there is no neuromechanical coupling effect between the commanded stiffness and force feedback that would result in a temporary loss of control over stiffness. This addresses R\ref{Requirements:5}.

\subsection{Visio-Verbal Teleimpedance Interface Architecture}
\label{sec:hardware_container_deployment}

\begin{figure*}[t!]
    \centering
    \includegraphics[width=1\linewidth]{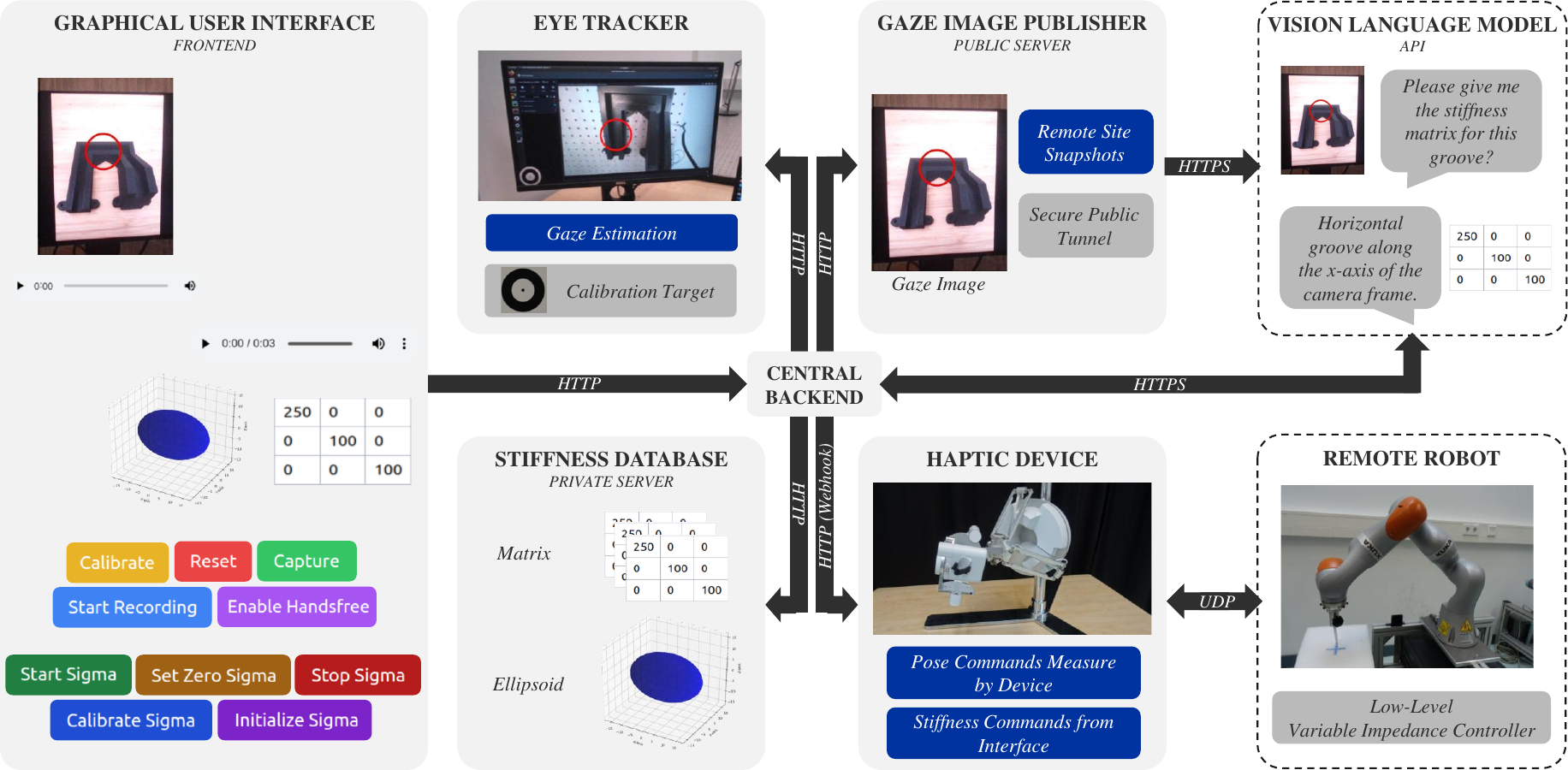}
    \caption{The diagram shows the container-based software architecture of the visio-verbal teleimpedance interface, emphasizing the communication links between the different containers. The Arrows indicate the data flow between these containers, and the blue blocks indicate the information that is passed. The blocks surrounded by a dotted line are not containers.}
    \label{fig:ConnectionDiagram}
\end{figure*}

The primary objective in designing the architecture of the proposed visio-verbal teleimpedance interface is to ensure the seamless integration of all its components while maintaining modularity and interchangeability. This flexibility allows the system to adapt easily to different hardware configurations and future updates. To accomplish this, we adopted a container-based approach, allowing the software modules to operate in isolated environments, whether remotely or locally, while preventing version conflicts. This design minimises the setup time and technical overhead, in line with the requirement for minimal configuration (R\ref{Requirements:4}). Figure \ref{fig:ConnectionDiagram} shows all the containers used in this interface and the connections between them.

Some of these containers are responsible for controlling the devices that the operator interacts with, such as the eye tracker, the haptic device or the graphical user interface (GUI). Each device is managed by a different container. The GUI is used for calibration, image capture, audio recording, and device status. It serves as the first point of entry for the operator. All requests from the GUI are sent to the \textit{Central Backend}, which processes and forwards them to the appropriate destination. The \textit{Central Backend} manages all the information flow, taking the information from all the services. 

On one side, we have the \textit{Eye Tracker} container that is responsible for connecting to the eye-tracking device, whether through a cable or wirelessly. It manages the short initial eye-tracking calibration process, performs gaze estimation, and overlays the estimated gaze point onto a snapshot of the operator's view. As a private server, there is also the \textit{Stiffness Database} that stores and provides generated stiffness matrices and ellipsoids, making them accessible to the \textit{Central Backend}. 

This information is collected by the \textit{Central Backend} to create a conversation history for the VLM. It also offers speech-to-text and text-to-speech capabilities, which are utilized in the process. The request is then sent to the Vision Language Model API. Since our VLM operates on external public servers, the \textit{Gaze Image Publisher} container takes snapshots of the remote site overlaid with gaze estimates represented as red circles, and with an internet-accessible tunnel allows the online VLM to retrieve gaze images through web-accessible URLs. The stiffness received from the VLM is then sent to the GUI by the \textit{Central Backend} to visually inform the operator of the current impedance configuration.

Finally, the stiffness is sent to the \textit{Haptic Device}. This container manages the haptic interface and ensures seamless communication with the remote robot controller via UDP communication, handling both pose commands and real-time stiffness updates. To improve safety, it includes a start/stop control mechanism that prevents unintended robot movements when the operator is not actively engaged. Additionally, using the GUI, the operator can automatically align the haptic device’s zero position with the robot’s current position, avoiding sudden jumps and ensuring smooth operation. This also enables workspace re-indexing during the operation.

\section{Experiments}
\label{sec:experiments}
\begin{figure}[t!]
    \centering
    \includegraphics[width=1\linewidth]{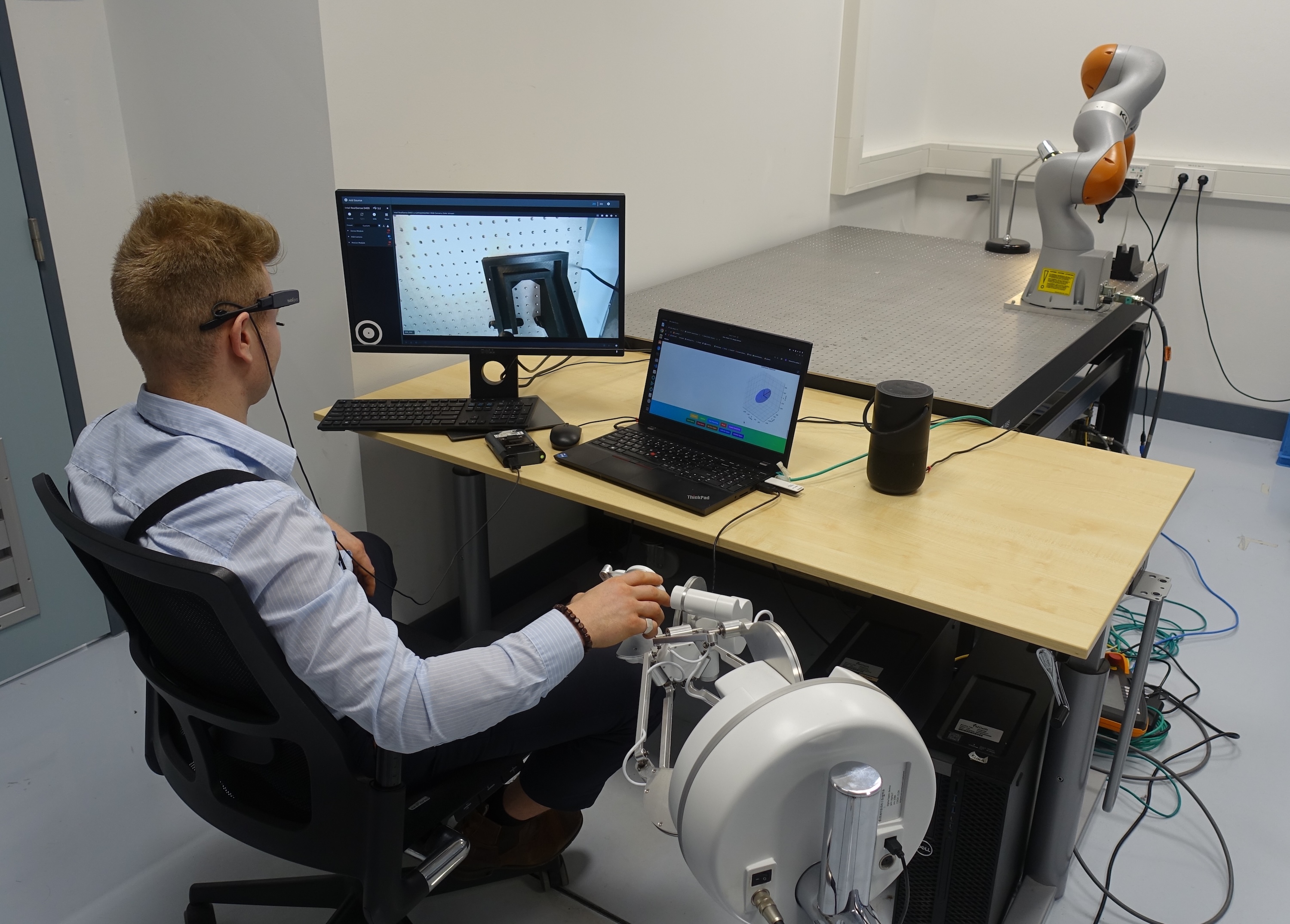}
    \caption{Experiment setup with teleimpedance system. The left screen provides a real-time camera feed mounted on the remote robot end-effector. The operator wears eye-tracking glasses and operates a haptic device to control the motion of the remote robot. The visio-verbal interface takes estimated gaze and verbal interaction as input and outputs the commanded stiffness for the remote robot based on the given task context.}
    \label{fig:hardware}
\end{figure}

To validate the proposed interface and demonstrate the key functionalities, we conducted experiments on a real teleoperation setup (see Fig.~\ref{fig:hardware}). At the local site, the operator monitors the remote environment via a display screen, which shows the live feed from a camera mounted at the end-effector of the remote Kuka LBR iiwa robotic arm. The operator commands the remote robot reference position through a Force Dimension Sigma.7 haptic device. The impedance adjustments are determined based on the operator’s gaze, recorded by an eye-tracking system (Tobii Pro Glasses 2), and the operator’s verbal interaction, captured by the built-in microphone on the laptop. The laptop also serves as the central processing unit for gaze and verbal input. The containers that run on it are created with \textit{Docker}, and the secure public tunnel to make the images available through public URLs is done with \textit{ngrok}. Finally, the VLM used is GPT-4o. To ensure transparency and confirm successful impedance adjustments, the system provides multimodal feedback. The selected stiffness matrix is conveyed through both a verbal notification via the laptop’s speakers and a visual representation in the form of a stiffness ellipsoid displayed on the GUI.

\begin{figure}[t!]
    \centering
    \includegraphics[width=0.75\linewidth]{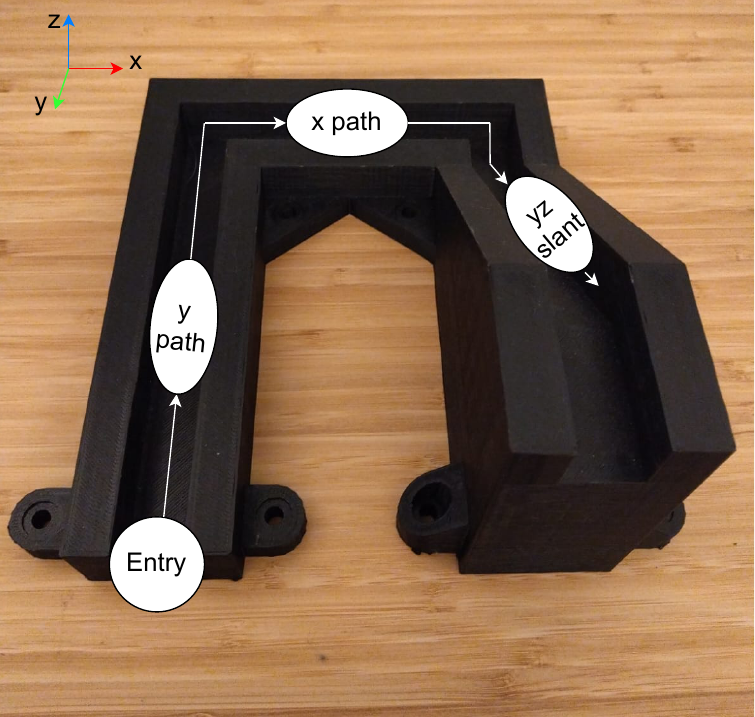}
    \caption{Groove structure with ellipsoid shapes (white) indicating the desired stiffness matrices for each phase of the task.}
    \label{fig:groove_structure}
\end{figure}

For the experiment task, we selected a \textit{slide-in-the-groove} task, which is common in assembly and involves physical interaction in multiple axes. We used a 3D-printed structure as seen in Fig. \ref{fig:groove_structure}, where the robot had to insert a peg into the groove and then navigate it while maintaining low interaction forces. The ability to dynamically adjust stiffness is critical to ensure smooth insertion, and subsequent traversing through the grooves without damage to the structure or the peg.

To perform this task optimally, the stiffness properties must be adapted across different phases of the task. At the entrance of the groove, the robot arm must be lowered down into the groove and align itself with the groove entrance without exerting excessive force on the structure. This is achieved by commanding a low stiffness in the x-axis and y-axis, allowing compliant self-alignment in that plane, while commanding \textit{high stiffness} in the z-axis to have the strength to push the peg inside. Once inside, the stiffness in the direction of the groove (y-axis) must be increased to ensure friction is compensated for accurate tracking of the reference trajectory. Meanwhile, stiffness in the directions of walls and bottom (x-axis and z-axis) should be low for the peg to comply with the environmental constraints and prevent excessive normal forces that could lead to jamming or structural damage. When the groove changes direction, the robot stiffness ellipsoid should be adjusted in a similar manner.

The experimental work was divided into several experiments. In the first experiment, we performed prompt optimization for the interface to explore different parameter settings and find the best one to be used for subsequent demonstration experiments. The purpose of the second experiment was to show verbal aspects of the interface focusing on demonstrating fluid task execution. The third experiment focused on demonstrating the gaze and conversational aspects.

\subsection{Experiment 1: Prompt Parameter Optimization}
The goal of the prompt parameter optimization study, specific to the use of VLMs in teleimpedance, was to investigate and compare the effects of different settings (system role, amount of priors, and image resolution) on the stiffness matrix predictions. VLMs extend the capabilities of large language models (LLMs) by incorporating image processing alongside textual input. A key discovery was their ability to perform few-shot learning, where instead of fine-tuning a large model for a specific task, a small set of example question-answer pairs alongside the user’s prompt are provided~\cite{brown2020fewshotlearning}. Studies have shown that few-shot prompting significantly improves VLM performance compared to zero-shot prompting, in which the model is asked to complete a task without prior examples~\cite{alayrac2022flamingo}.

Our research investigates an extension of few-shot learning in VLMs by incorporating gaze-estimate images alongside a curated set of labelled examples that define the corresponding stiffness matrices. This approach leverages the few-shot paradigm to enhance the model’s ability to infer appropriate stiffness configurations based on visual and verbal contextual cues. The way the VLM is prompted significantly influences the quality of its outputs. To ensure optimal performance by generating accurate stiffness matrices in different phases of the task, we conduct a parameter optimization experiment. This allowed us to identify the most effective combination of prompt parameters for teleimpedance application, including the task description, example demonstrations, and image detail.

\begin{table*}[t]
    \centering
    \renewcommand{\arraystretch}{1.2}
    \begin{tabular}{|c|c|c|c|c|c|c|c|c|}
        \hline
        Role & Prior & Resolution & Entrance & Y-traverse & X-traverse & YZ Slant & Overall (With Slant) & Overall (No Slant) \\
        \hline
        Role 3 & Lab  & High & 1.00 $\pm$ 0.00 & 0.93 $\pm$ 0.06 & 1.00 $\pm$ 0.00 & 0.00 $\pm$ 0.00 & 0.73 $\pm$ 0.06 & 0.98 $\pm$ 0.02 \\
        Role 1 & Lab  & High & 1.00 $\pm$ 0.00 & 0.67 $\pm$ 0.12 & 0.93 $\pm$ 0.06 & 0.13 $\pm$ 0.09 & 0.68 $\pm$ 0.06 & 0.87 $\pm$ 0.05 \\
        Role 2 & Lab  & High & 0.93 $\pm$ 0.06 & 0.67 $\pm$ 0.12 & 1.00 $\pm$ 0.00 & 0.07 $\pm$ 0.06 & 0.67 $\pm$ 0.06 & 0.87 $\pm$ 0.05 \\
        Role 3 & Ideal & High & 0.93 $\pm$ 0.06 & 0.67 $\pm$ 0.12 & 0.93 $\pm$ 0.06 & 0.00 $\pm$ 0.00 & 0.63 $\pm$ 0.06 & 0.84 $\pm$ 0.05 \\
        Role 1 & Lab  & Low  & 1.00 $\pm$ 0.00 & 0.27 $\pm$ 0.11 & 1.00 $\pm$ 0.00 & 0.00 $\pm$ 0.00 & 0.57 $\pm$ 0.06 & 0.76 $\pm$ 0.06 \\
        Role 2 & Lab  & Low  & 1.00 $\pm$ 0.00 & 0.00 $\pm$ 0.00 & 1.00 $\pm$ 0.00 & 0.00 $\pm$ 0.00 & 0.50 $\pm$ 0.06 & 0.67 $\pm$ 0.07 \\
        Role 3 & Lab  & Low  & 1.00 $\pm$ 0.00 & 0.00 $\pm$ 0.00 & 1.00 $\pm$ 0.00 & 0.00 $\pm$ 0.00 & 0.50 $\pm$ 0.06 & 0.67 $\pm$ 0.07 \\
        Role 1 & Ideal & High & 0.73 $\pm$ 0.11 & 0.40 $\pm$ 0.13 & 0.60 $\pm$ 0.13 & 0.00 $\pm$ 0.00 & 0.43 $\pm$ 0.06 & 0.58 $\pm$ 0.07 \\
        \hline
    \end{tabular}
    \caption{Mean prediction accuracy values with standard deviations (mean $\pm$ std) for the main test.}
    \label{tab:exp1}
\end{table*}

In deep learning terminology, the \textit{system role} directs the VLM to a specific function, which was, in our case, a specialized stiffness matrix generator. This ensured that operator input was interpreted in terms of matrix updates rather than just general conversation. We evaluated three prompt designs (system roles) with progressively stronger inductive bias, each building on the previous one. Role 1 provides only an instruction to generate a stiffness matrix from the most recent input image, the required response format, and permission to use conversation history (i.e., few-shot examples of image–stiffness matrix pairs). Role 2 introduces the task physics as an impedance-control problem (virtual spring aligned with the groove), adds a solution procedure, and specifies the camera frame and numeric guidance (e.g., compute stiffness for the highlighted groove; prescribe high stiffness 250 along the groove and low stiffness 100 in orthogonal directions and express it in terms of the Cartesian camera frame). Building on Role 2, Role 3 further includes textual labels for specific groove sections with their corresponding target stiffness matrices (e.g., “groove along x-axis (left–right)” corresponds to diag(250, 100, 100)).

Another important aspect is related to the few-shot demonstration approach and the inclusion of conversation \textit{priors}. These contain example prompts (e.g., ``What is the stiffness matrix for this phase?'') with corresponding images for different phases of the task paired with the desired outputs in the form of stiffness matrices for the given phases of the task. Since we had four phases of the task (entrance, y-traverse, x-traverse, and yz-slant), four example prompts and four corresponding images were fed to the VLM. This list was tested under three conditions. In the None condition, the model was tested without any example prompts and responses. In the Ideal condition, the prior message list was created from a setup in the environment, which had less challenging lighting and camera angle than the lab environment. In the Lab condition, the prior message list was created from the lab environment, which had challenging lighting and camera angle. The inclusion of the Ideal condition allowed us to investigate whether the model's performance would improve when provided with prior data from either the ideal or lab environment, shedding light on how environmental differences influence the results.

Finally, the image detail was varied between low and high settings to determine its impact on performance. The \textit{low-detail} mode provided faster responses, while the \textit{high-detail} mode allowed the model to process finer details in the input images. In the low-detail mode we down-sampled the image to $512 \times 512$ and budgeted 85 tokens for image description, while in the high-detail mode, we allocated more tokens (e.g., 170 per tile) to achieve detailed image analysis, improving the model’s ability to extract relevant gaze-based features.

The conversation history maintains a record of the operator’s previous commands and context, including gaze snapshots, enabling iterative refinement of stiffness matrices. For the tests in Experiment 1, we had only priors and no additional conversation history to make sure that the comparison between conditions was fair. In the subsequent demonstration experiments (i.e., 2 and 3), we let the VLM keep the conversation history to continue improving and learning. Since more history also makes VLM slower, we limited it to 10 prompts to ensure faster responses.

\begin{figure}[t!]
    \centering
    \includegraphics[width=1\linewidth]{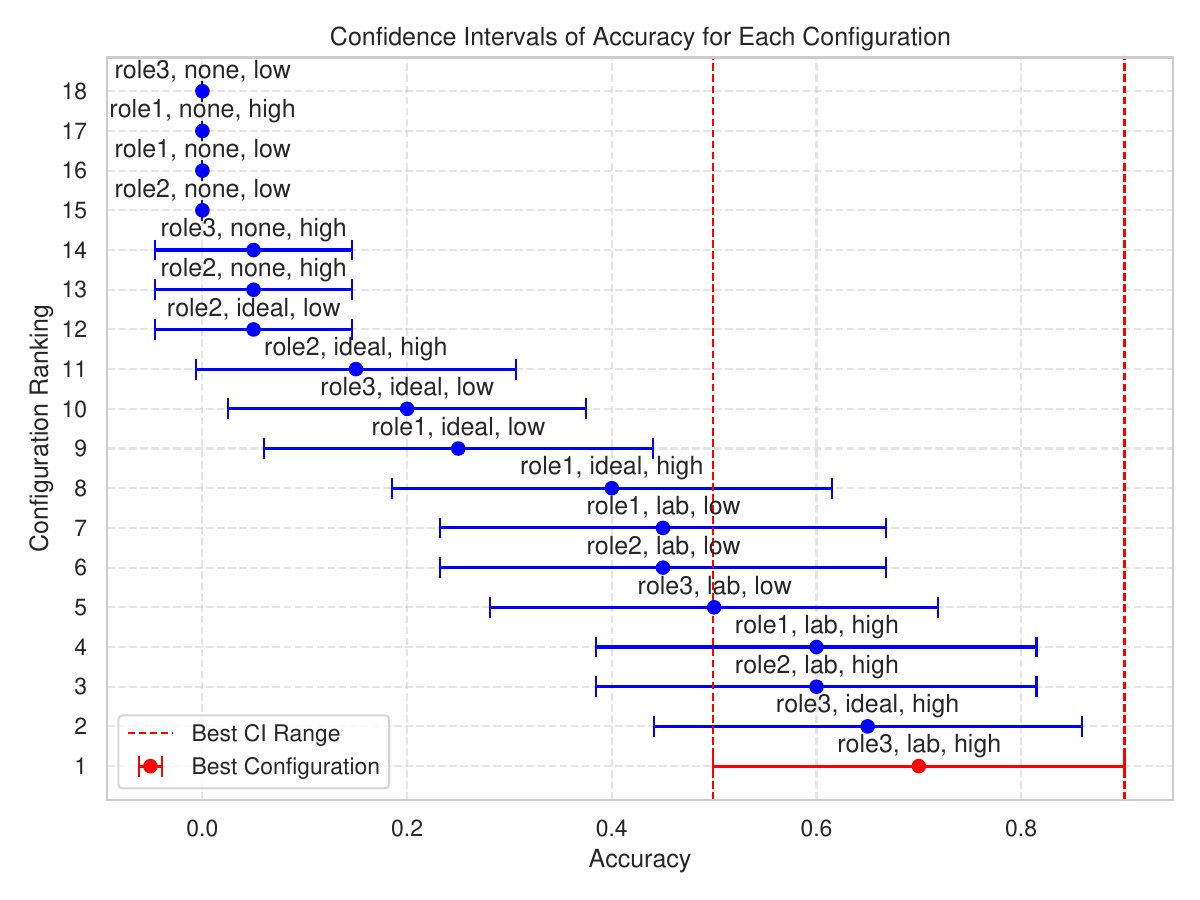}
    \caption{Results of a pre-test for all prompt configurations with means (dots) and confidence intervals (whiskers). The best nine configurations from the pre-test were used for the main test on more trials in Tab. \ref{tab:exp1}.}
    \label{fig:exp1}
\end{figure}

We first performed a pre-test on the accuracy of all possible combinations with 5 trials for each combination to eliminate the worst-performing ones. In each trial, we tested whether the VLM could correctly generate the stiffness matrix for the four different phases of the task (entrance, y-traverse, x-traverse, and yz-slant). Figure \ref{fig:exp1} shows the accuracy results of the pre-test for the four different phases of the task for all possible combinations. The accuracy was calculated as the percentage of correct stiffness predictions out of the total trials. The correct stiffness was defined based on the optimal strategy for a given phase of the task, as previously defined above. For example, during sliding, the correct stiffness was defined to be high in the direction of pushing, and low stiffness in the direction to the walls.

For the experiment, we selected a challenging scene with difficult lighting and camera angle in the lab environment. Note that since the camera angle was directly above the structure, it was impossible for the vision to identify the upward slant, the results for the slat phase are accordingly poor. Nevertheless, for the other phases of the task where vision could detect the structure, VLM achieved excellent accuracy.

Following the pre-test, we selected the nine best-performing combinations for more extensive testing, where we used 15 trials for each combination. We kept the setting from the pre-test with a challenging scene with difficult lighting and camera angle. Table \ref{tab:exp1} shows the results of the main test with the average accuracy based on entrance, y-traverse, x-traverse, and yz slant phases. Based on the insight from this experiment, we determined that the optimal prompt configuration was: an elaborate task description with labelled examples (Role 3) for the system role, the inclusion of lab-based exemplars as a primer (Lab), and high-detail image processing (High). This refined configuration was then used in the subsequent demonstration experiments.

\subsection{Experiment 2: Demonstration of Verbal Commands}
This experiment aimed to demonstrate the interface in a case when the operator wants a fluid task execution with minimum interaction with the interface. The operator issues purely verbal commands (e.g., ``Increase stiffness along the groove axis'') without providing gaze snapshots. This demonstration verifies that the interface can still produce meaningful matrices when images are absent, although it relies on user-specified axis names rather than visual context.

\begin{figure}[t!]
    \centering
    \includegraphics[width=1\linewidth]{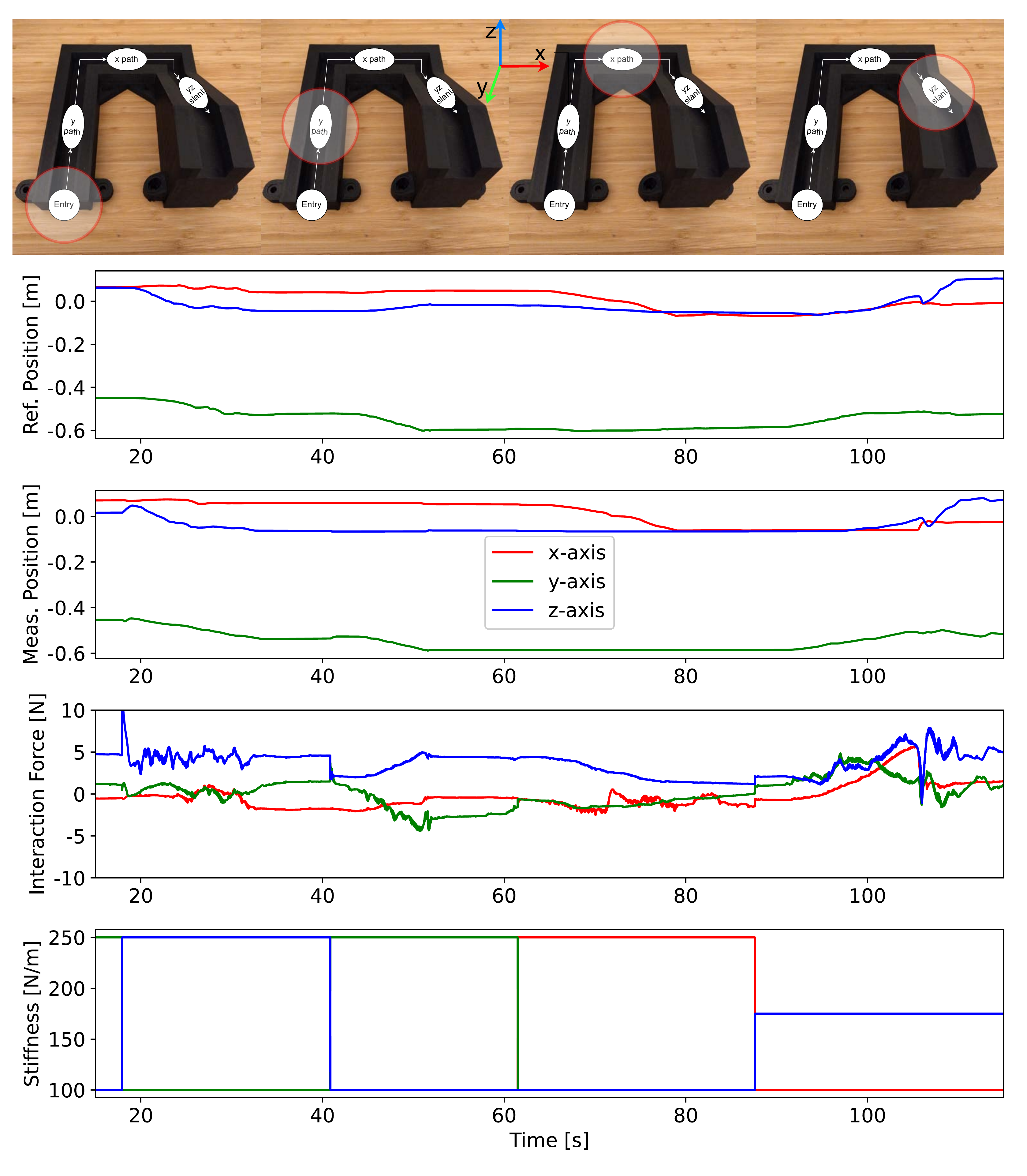}
    \caption{Results of the experiment with swift verbal commands. The images on top show key phases of the task in the groove structure, including entrance, y-traverse, x-traverse and yz slant. The graphs show reference (first) and measured (second) robot end-effector position, measured forces (third), and stiffness (fourth). Different colours indicate variables in different axes. The stiffness changes also indicate transitions between phases.}
    \label{fig:exp2}
\end{figure}

The results are presented in Fig. \ref{fig:exp2}. We can see the different phases of the task from the series of images on top. The robot reference movement as commanded by the operator is seen in the first graph, and the corresponding actual movement of the robot is shown in the second graph. The fourth graph shows the robot stiffness adjustments by the interface based on the operator's verbal command.

When the operator informed the interface about the intention to enter the groove, the stiffness was adjusted in a way that the robot became stiff along the z-axis and compliant along the other axes. This corresponds to a standard peg-in-the-hole strategy, where the robot should be stiff in the direction of pushing and compliant elsewhere to allow the peg to comply with the environment and slide inside. By observing the blue line at around 20 seconds, we can see how the operator then performed the insertion of the peg into the groove along the vertical direction (z-axis).

In the next phase, the operator communicated to the interface the intention to traverse along the y-axis inside the groove. The stiffness was changed so that the robot became stiff along the y-axis and compliant along the other axes. This ensured that the peg overcame the friction needed to move in the y-axis, while interaction forces with the rigid wall were kept low. Green lines on the first and second graphs at around 45 seconds show the robot movement that the operator performed along the groove in the y-axis.

A similar process can be observed for movement in the groove along the x-axis around 65 seconds, where stiffness was changed to be high in the x-axis. In the final phase, the movement in the groove was slanted upward and when the operator informed the interface about the intention, the stiffness was adjusted to be stiff along 45 degrees in the y-z plane. The final slanted movement of the robot can be seen around 100 seconds.

The third graph shows that the interaction force on average stayed below 5 N, with only one short peak around 10 N. This interaction force performance is comparable to a study in the literature, where a hand-held manual teleimpedance interface was used for the robot control to navigate the same structure~\cite{kraakman2024design}.

\subsection{Experiment 3: Demonstration Gaze and Conversational Aspects}
This experiment aimed to demonstrate the interface with combined gaze and conversational aspects. While the focus of the previous experiment was on fluid task execution with the operator providing swift verbal commands, here the focus is on rich visio-verbal interaction. The gaze estimate and scene snapshot are included, allowing the operator to state the intentions (e.g., ``I want to enter the structure''), while the VLM uses both the gaze-labeled snapshot and voice transcription to generate a more contextualized matrix.

\begin{figure}[t!]
    \centering
    \includegraphics[width=1\linewidth]{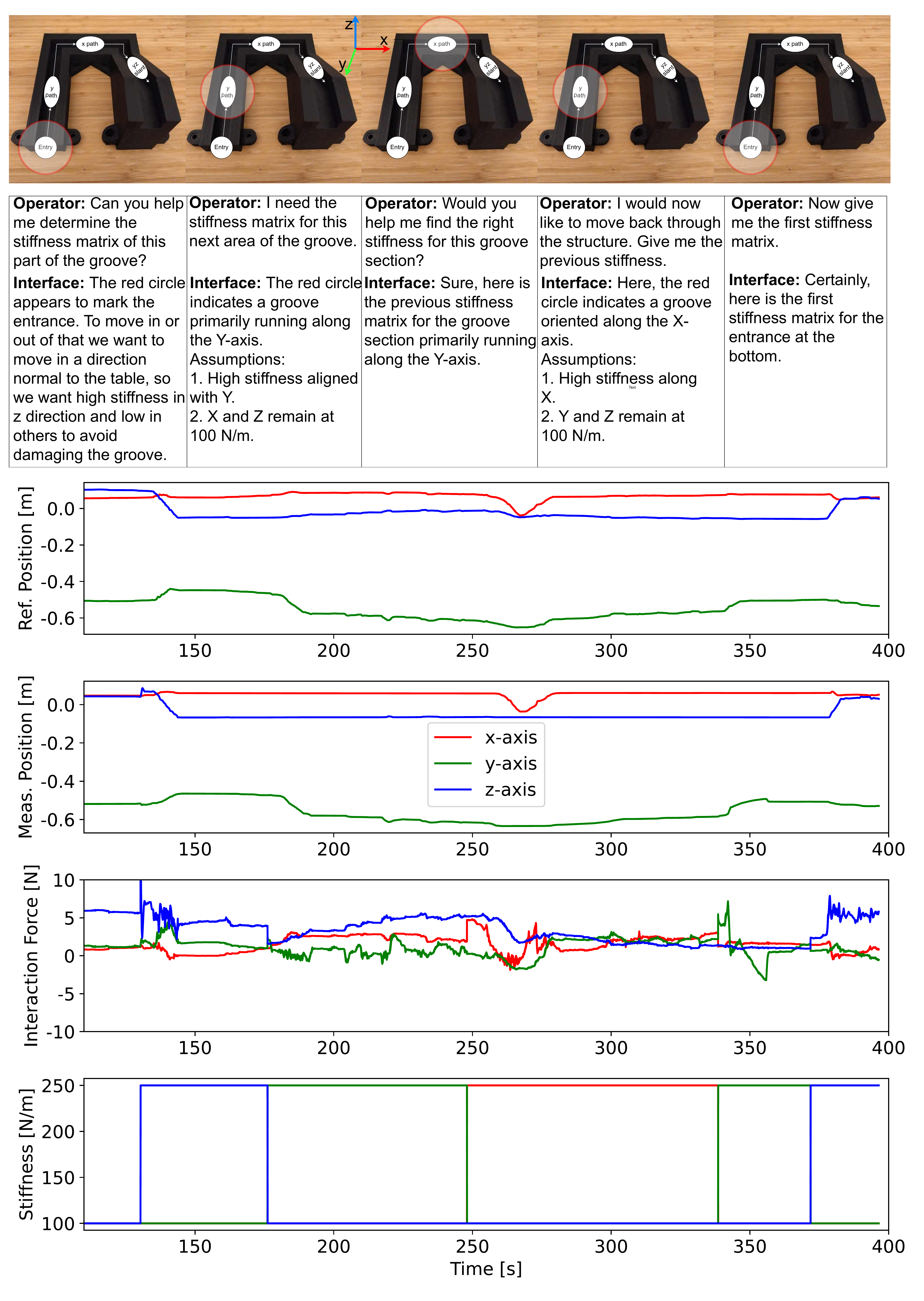}
    \caption{Results of the experiment with combined gaze and conversational aspects. The layout is the same as Fig. \ref{fig:exp2}. Additionally, it lists the conversation history between the operator and the interface in each phase of the task. The stiffness changes also indicate transitions between phases.}
    \label{fig:exp3}
\end{figure}

The results are presented in Fig. \ref{fig:exp3}. The different phases of the task are illustrated with a series of images on top. The graphs show the same variables as in the previous experiment. The conversational history between the operator and the interface in each phase of the task is also presented in the figure. In the first phase, the operator looked at the entrance and communicated the intention to enter the structure to the interface verbally. Through the detected gaze and verbal context, the interface generated the correct stiffness matrix to enter the groove and gave the operator a verbal confirmation with some extra details. After the operator entered the groove with the peg on the remote robot end-effector, the gaze was changed to the next phase and a new verbal context related to the movement along that section was given to the interface. After the verbal confirmation, the operator moved along the y-axis. A similar procedure was done for the next phase, which involved movement along the x-axis.

At that point, the operator wanted to show the advantage of conversation history in the interface by reversing the movement to backtrack along the structure. Instead of communicating specific intentions regarding each phase, the operator just instructed the interface that he wanted to backtrack. The interface then sequentially changed the robot stiffness matrix when backtracking the previous phases until the operator exited the structure at the point it was entered. In case the interface made a mistake in classification and generated an incorrect stiffness matrix, the operator could use verbal communication to provide corrections.

\section{Discussion}
\label{sec:discussion}
The results show that the visio-verbal teleimpedance interface can generate adaptive stiffness matrices under dynamic verbal interaction and visual cues. Experiment 1 provided insights into prompt engineering that optimized for teleimpedance applications. Experiment 2 demonstrated the use case where the operator can use the proposed interface for fluid task execution with minimal visio-verbal interaction, at the expense of advanced functionalities. Experiment 3 showed the full potential of visio-verbal interaction with rich visio-verbal interaction, where the operator could naturally indicate things by gaze and also receive verbal confirmations.


The experiments also highlighted the limitation of using a single camera angle as an input into VLM. For complex structures like the one used in this study, some features may be difficult to extract by vision. For example, in a top-down camera view and poor lighting conditions, the slant in the structure can be impossible to recognise. In this camera angle, it was simply misclassified as y-traverse. We performed a post-experiment test with a slightly easier camera angle and better lighting conditions (i.e., the one seen in Fig. \ref{fig:groove_structure}). In that case, the classification accuracy for slant increases to 66.6\%. A promising future work would be to design an autonomous camera control method, where the robot would take images of the scene from multiple angles.

The prompt optimization experiment provided some additional insights regarding the teleimpedance application. For example, the image resolution has a major effect on the prediction accuracy. As seen in Tab. \ref{tab:exp1}, four top-performing configurations all have high-resolution images. Another interesting insight is that including image priors with very good lighting/camera conditions that are from a different environment (Ideal) than the one used in the actual task (Lab) does not help the performance of the teleimpedance interface. As seen in Tab. \ref{tab:exp1}, only two configurations in the nine top-performing had Ideal priors, and the top three all had Lab priors.

One potential limitation of this study is the reliance on a mobile eye tracker with glasses, which necessitates additional image-processing steps to isolate the display screen from the surrounding environment. An alternative would be to employ a static eye tracker on the display screen itself.

\section*{Acknowledgement}
{The authors would like to thank Renchi Zhang for the help and advice regarding the eye tracking system and Yke Bauke Eisma for lending the eye tracking system.}

\bibliographystyle{IEEEtran}
\small
\setstretch{0.0}
\bibliography{preprint}

\end{document}